\documentstyle[11pt,bigpage,epsfig]{article}
\title{Using single layer networks for discrete, sequential 
data: an example from natural language processing\\
\scriptsize 
Published in Neural Computing \& Applications, 1997, Vol 5 (4) }
\normalsize
\author{Caroline Lyon and Ray Frank\\
School of Information Sciences \\
University of Hertfordshire \\
Hatfield, Herts. AL10 9AB\\
{\tt email: C.M.Lyon@herts.ac.uk}~~~~Tel: +44 (0)1707 284266}
\date{}
\begin{document}
\maketitle
\begin{abstract}
Natural Language Processing (NLP) is concerned with processing ordinary,
unrestricted text. This work takes a new approach to a traditional NLP task,
using neural computing methods. A parser which has been successfully 
implemented is described. It is a hybrid system, in which neural 
processors operate within a rule based framework.

The neural processing components belong to the class of Generalized Single
Layer Networks (GSLN). In general, supervised, feed-forward networks need 
more than one layer to process data.  However, in some cases data can be 
pre-processed with a non-linear transformation, and then presented in a 
linearly separable form for subsequent processing 
by a single layer net. Such networks offer advantages of functional 
transparency and operational speed.

For our parser, the initial stage of processing maps linguistic data onto a 
higher order representation, which can then be analysed by a single layer 
network. This transformation is supported by information theoretic analysis.

Three different algorithms for the neural component were investigated.
Single layer nets can be trained by finding weight adjustments based on
(a) factors proportional to the input, as in the Perceptron, (b) factors 
proportional to the existing weights, and (c)   an error 
minimization method. 
In our experiments generalization ability varies little;
method (b) is used for a prototype parser. This is available via telnet.

\end{abstract}
 
\section{Introduction}

This paper examines some of the issues that have to be addressed in
designing neural processors for discrete, sequential data. There is a mutual 
dependence between the
representation of data, on the one hand, and the architecture
and function of an effective network on the other. 
As a vehicle for examining these processes we describe an automated partial
parser that has been successfully developed \cite{lyon6,lyon9}. This takes
natural language sentences and returns them with the subject and head of the
subject located.
Ability to generalize is the primary concern. A prototype can be accessed
via telnet, on which text can be entered and then parsed. Intermediate steps in
 the process can be seen.\footnote{For details, contact  author.}

In principle, simpler networks with well understood functions have
{\em prima facie} advantages, so looking for a representation that enables
such networks to be used should be advantageous. With
feed forward, supervised networks single layer models enjoy
functional transparency and operational speed,  but in general 
this type of network  will need  more than one dynamically linked layer to 
model non-linear relationships.

However, there is an alternative approach. The layers may be 
de-coupled, and processing at different layers done in separate steps.
Data can be transformed, which is analogous to processing at the first layer,
and then presented 
in a linearly separable form to a single layer net, which is analogous to
a second layer. This is illustrated in Figure~\ref{gsln2}, which shows 
in simplified form an archetype of the 
class of Generalized Single Layer Networks (GSLN). A
number of different network types that belong to this class are listed
by Holden and Rayner~\cite[page 369]{holden}. 
The critical issue is
finding  an appropriate non-linear transformation to convert data into 
the required form.

This paper describes 
how characteristic linguistic data can be  converted into
a linearly separable representation that partially captures sequential 
form. The transformed data is then processed by a single layer network. Three 
different neural models are tried, and their 
performance is compared. 

All three networks are feed forward models with supervised training. 
Connection weights can be found by adjustments based on
(a) factors proportional to the input (b) factors proportional to the 
existing weights, and (c)  factors related to the
 difference between desired and actual output, an error minimization
method. Model (a) is a 
traditional Perceptron; model (b) is based on the
Hodyne network introduced by Wyard and Nightingale at British
Telecom Laboratories~\cite{wyard}; model (c) comes from the class of networks
that use an LMS (Least Mean Square error) training algorithm.
There is little difference in generalization ability, but
network (b) performs slightly better and has been used for the parser in the 
prototype.

\subsection*{Natural language processing (NLP)}
The automatic parsing of
natural language poses a significant problem, and neural computing techniques
can contribute to its solution. For an overview of the scope for  work 
in NLP see \cite[pages 4-11]{cunningham}.
Our prototype gives results of over 90\% correct on declarative
sentences from technical manuals (see Section~\ref{results}).

Automated syntactic analysis of natural language has, in the last decade, 
been characterised by two paradigms. Traditional AI, rule based methods 
contrast with probabilistic
approaches, in which stochastic models are developed from large corpora of
real texts. Neural techniques fall into the broad category of the latter, data
driven methods, with
trainable models developed from examples of known parses. 
The parser we have implemented uses a hybrid approach: rule based 
techniques are integrated with neural processors.
  
Parsing can be taken as a pattern matching
task, in which a number of parses are postulated for some text. A classifier 
distinguishes between the desired parse  
and incorrect structures. The pattern matching capabilities of neural networks
 have a particular contribution to make to the 
process, since they can conveniently model negative as well as positive 
relationships. The occurrence of some words or groups of words inhibit
others from following, and these constraints can be exploited.
 Arguments on the need for negative 
information in processing formal languages \cite{gold1} can be 
extended to natural language. 
This is an important source of information 
which has been difficult for traditional probabilistic 
methods to access~\cite{charniak2,brill}.
Neural methods also have the advantage that 
training is done in advance, so the run time computational load is low.

\subsection*{Contents of paper}
This paper will first take an overview of some factors that are relevant to 
neural net design decisions, (Section~\ref{classifiers}). 
It then looks at  characteristics of natural language 
(Section~\ref{natlang}), and the representation of 
sequential data (Section~\ref{seq-dat}). 
A description of the hybrid system used in our work is given, 
(Section~\ref{proto}).
Then we examine some of the design issues for  
the neural components of this system. First, the data itself is 
examined closely. Then we consider how the data
can be transformed
for processing with a single layer net (Section~\ref{lin-sep}).
 We also comment on the use of a Bayesian classifier, which performs
slightly less well than the neural networks. 
Section~\ref{single} describes
 the three different networks: (a) the Perceptron, (b) Hodyne and
(c) an LMS model.

In Section~\ref{results} we compare the performance 
of the three networks. Generalization is good, 
providing that enough  training  data is used.  Over
90\% of the data is correctly classified, and the output can be interpreted 
so that results for the practical application are up to 100\% correct.
On the small amount of data processed so far the different networks have 
roughly comparable generalization ability, but the Hodyne model is slightly
better. A discussion on the function of the net follows in 
Section~\ref{op-of-net}.
We conclude (Section~\ref{conclusion}) that linguistic data is 
a suitable candidate for processing with this approach.

\section{Neural nets as classifiers of different types of data}
\label{classifiers}
\subsection{``Clean'' and noisy data}
Consider the
fundamental difference in purpose between systems that
handle noisy data, where it is desired to capture an underlying structure 
and smooth over some input, compared to those that
process ``clean'' data, where every input datum may count. 
The many applications
of neural nets in areas such as image processing provide examples of the 
first type,
the parity problem is typical of the second.\footnote{The classical 
parity problem takes a binary input vector, the elements of which are $0$ or 
$1$, and classifies it as having an even or odd number of $1$'s.} 
These ``clean'' and ``noisy'' types can be 
considered as  endpoints of a spectrum, along which different
processing tasks lie. In the case of noisy data
a classifier will be required to model significant characteristics in
order to generalize effectively. The aim is to model the underlying function
that generates the data, so the training data should not be over fitted.

On the other hand, for types of data such as inputs to a parity
detector, no datum is noise. 
Consider an input pattern that is 
markedly different, that is topologically distant, from others in its class. 
For one type of data this may be noise. In other instances an
``atypical" vector may not be noise, and we may need to capture the
information it carries to fix the class boundary effectively.

As we demonstrate in the next section, linguistic data needs to be 
analysed from both angles. We need to capture the statistical information
on probable and improbable sequences of words; we also need to use the
information from uncommon exemplars, which make up a very large proportion 
of natural language data. 

\subsection{Preserving topological relationships}
\label{topological}
Another of the characteristics that is relevant to  network design is the
extent to which the classification task, the mapping from input to
output, preserves topological relationships. In many cases data which are
close in input space are likely to produce the same output, and conversely
similar classifications are likely to be derived from similar inputs. 
However, there are other classification
problems which are different: a very small change in input may cause a
significant change in output and, on the other hand, very different input
patterns can belong to the same class. Again, the parity problem is a 
paradigm example: in every case changing a single bit in the input pattern 
changes the desired output.

\subsection{Data distribution and structure}
\label{distrib}
Underlying data distribution and structure have their effect on the 
appropriate type of processor, and these characteristics should be examined.
Information about the 
structure of linguistic data can be used to make decisions on suitable 
representations. In this work information theoretic techniques are
used to support decisions on representation of linguistic data.

We may also use information on
data distribution to improve generalization ability. As shown in 
Section~\ref{natlang}, assumptions of
normality cannot be made for linguistic data. 
The distribution indicates that in order to
generalize adequately the processor must capture information from a
significant number of infrequent events.

\section{Characteristics of linguistic data}
\label{natlang}
The  significant characteristics of natural language that we wish to
capture include:
\begin{itemize}
\item An indefinitely large vocabulary
\item The distinctive distribution of words and other linguistic data
\item A hierarchical syntactic structure
\item Both local and distant dependencies, such as feature agreement
\end{itemize}

\subsection{Vocabulary size}
Shakespeare is said to have used a vocabulary of 30,000 words, and even an
inarticulate computer scientist might need 15,000 to get by  (counting 
different forms of the same word stem separately).
Current vocabularies for commercial speech processing databases 
are  $O(10^5)$. Without
specifying an upper limit we wish to be able to model an indefinite number
of words.

\subsection{The distinctive distribution of linguistic data}
The distribution of words in English and other languages has
distinctive characteristics, to which Shannon drew attention \cite{shannon}.
Statistical studies were made of word frequencies in English language texts.
In about 250,000 words of text (word tokens) there were about 30,000 different
words (word types), and the frequency of occurrence of each word type was
recorded. If word types are
ranked in order of frequency, and $n$ denotes rank,  then there is an empirical
relationship between the probability of the word at rank $n$ occurring,
$p(n)$, and $n$ itself, known as Zipf's Law: 
\[              p(n) \ast n = constant       \] 

This gives a surprisingly good approximation to word probabilities in
English and other languages, and indicates the extent
to which a significant number of words occur infrequently.
For example, words that have a frequency of less than 1 in 50,000 make up 
about 20-30\% of typical English language news-wire reports \cite{dunning}.
The LOB corpus\footnote{The London Oslo Bergen corpus
is a collection of texts used as raw material for natural language 
processing} with about 1 million word tokens  contains about 50,000 different
word types, of which about 42,000 occur less than 10 times each
\cite{atwell4}.

The ``zipfian'' distribution of words has been found typical of 
other linguistic data. It is found again in the data 
derived from part-of-speech tags used to train the prototype described 
here: see Figures~\ref{graph_a}~and~\ref{graph_b}. 

Other fields in which zipfian distribution is noted include information 
retrieval and data mining (e.g. characteristics of WWW use, patterns of
database queries). It has also been observed in molecular biology
(e.g. statistical characteristics of RNA landscapes, DNA sequence coding).

\subsubsection*{Mapping words onto part-of-speech tags}
In order to address the problem of sparse data 
the vocabulary can be partitioned into
groups, based on a similarity criterion, as is done in our system.
 An indefinitely large
vocabulary is mapped onto a limited number of part-of-speech tag classes.
This also make syntactic patterns more pronounced. Devising optimal
tagsets is a significant task, on which further work remains to be done.
For the purpose of this paper we take as given the tagsets used in the 
demonstration prototype, described in \cite{lyon-thesis}. At the stage of
processing described in this paper 19 tags are used.

\subsection{Grammatical structure}

\begin{figure}[hbt]
\begin{center}
\epsfig{figure=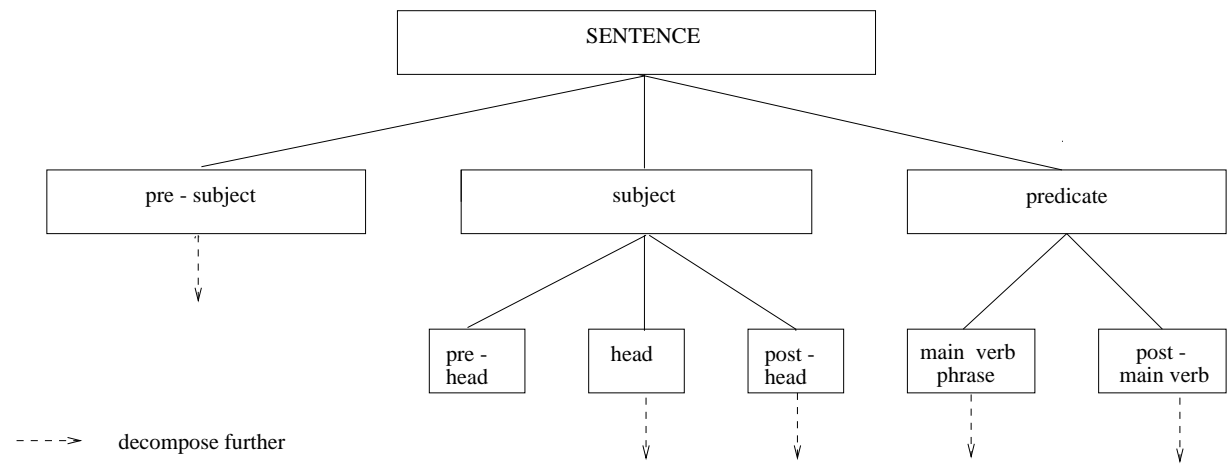,width=\textwidth}
\end{center}
\caption{Decomposition of the sentence into syntactic constituents.
\label{harchy}}
\end{figure}

There is an underlying hierarchical structure to all natural languages,
a phenomenon that has been extensively explored. Sentences will usually 
conform to certain structural patterns, as is shown in a simplified form in 
Figure \ref{harchy}. 
This is not inconsistent 
with the fact that acceptable grammatical forms evolve with time, and
that people do not always express themselves grammatically.
Text also, of course, contains non-sentential elements such as headings,
captions, tables of contents. The work described in this paper is restricted
to declarative sentences. 

Within the grammatical structure there is
an indefinite amount of variation in the way in which words can be
assembled. On the other hand, the absence
or presence of a single  word  
can make a sentence unacceptable, for example
\begin{quote}
$\ast$There are many problems arise. \quad\ldots(1)\\
$\ast$We is late. \quad\ldots(2)
\end{quote}

Now consider how linguistic data fits into the scheme described in 
Section \ref{topological} on preserving topological relationships.
Many strings of words that are close in input space are also in the same
grammatical category, but, conversely, on occasions a single word change 
can put a string
into a different category. Our processor has to model this.
  
\subsection{Local and distant dependencies}

In examining natural language we find there are dependencies between certain
words, both locally and at a distance. For instance, a plural subject 
must have a plural verb, so a
construction like sentence (2) above is incorrect. This type of dependency 
in its general form is not necessarily local.
In sentence (3) below the number of the subject is determined by its head, 
which is not adjacent to the verb:
\begin{quote}
If a cooler is fitted to the gearbox, [~~the pipe~~[~~connections~~] of the
cooler~~] must be regularly checked for corrosion. \quad\ldots(3)
\end{quote}
\noindent

The subject of this sentence is the plural ``connections''. Note that modal 
verbs like ``must'' have the 
same singular and plural form in English, but not in many other languages. 
For an automated translation system to process modal verbs it
is necessary to find the head of the subject that governs the verb and ensure
number agreement.

There are also dependencies between sentences and between more distant
parts of a text.
We aim to model just the intra-sentential dependencies as our automatic
parser is developed. 

\section{Modelling sequential data} 
\label{seq-dat}
Three methods have commonly been used to model sequential data, such as 
language, for connectionist processing. The first is to
move a window along through the sequence, and process a series of 
static ``snapshots''. Within each window ordering information is not 
represented.  Sejnowski's NETtalk is a well known 
example~\cite{sejnowski}.

Another method that warrants further investigation
is the use of recurrent nets \cite{elman2,giles1}. In its basic form this 
type of network is 
equivalent to a finite state automaton that can model regular 
languages \cite{giles3}.

\subsection{The n-gram method}
\label{tuples}
The third method, used in this work, is to take sets of ordered, 
adjacent elements, which capture some of the sequential structure of
language. This is related to the well known trigram approach used in
probabilistic language processing.
Combining tags into higher order tuples can also
act as a pre-processing function, making it more likely that  the transformed 
data can be processed by a single layer network (Section~\ref{lin-sep}).  

This method of representation captures some of the structure of natural
language, as is shown by analysis with information theoretic techniques.
There are relationships between neighbouring words in text:
some are likely to be found adjacent, others are unlikely. When words are 
mapped onto part-of-speech tags this is also the case.  
This observation is 
supported by an investigation of entropy levels in the LOB corpus, in 
which 1 million words have  been manually tagged. 

Entropy can be understood as a measure of uncertainty 
\cite[chapter 2]{cover2}. The uncertainty about how a partial sequence will
continue can be reduced when statistical constraints of neighbouring elements
 are taken into account. 
Shannon introduced this approach by analysing sequences of letters 
\cite{shannon}, where the
elements of a sequence are single letters, adjacent pairs or triples, 
with order preserved. These 
are n-grams, with $n$ equal to 1, 2 or 3. 
The entropy of a sequence represented by letter n-grams declines as $n$
increases. When  sequences of tags in the LOB corpus were analysed
the same result was obtained: the entropy of part-of-speech n-grams declines 
as $n$ increases from 1 to 3.
This indicates that some of the structure of language is
captured by taking tag pairs and triples as processing elements. 

We adopt the common approach of presenting data as  binary vectors for all
the networks examined in this work. Each element of the input vector 
represents an ordered tuple of adjacent part-of-speech tags,
a pair or a triple. If a given tag tuple is present in an input string, then 
that element in the input vector is flagged to 1, else it remains 0. 

\section{Description of the hybrid natural language processor}
\label{proto}

In order to process unrestricted natural language it is 
necessary to attack the problem on a broad front, and use every possible
source of information. In our work the neural networks are part of a larger 
system, integrated with rule based modules. We first {\em assert} that there 
is a syntactic structure which can be mapped onto a sentence 
(Figure~\ref{harchy}). Then we use neural methods to find the mapping in each 
particular case. The grammar used is defined in \cite[chapter 5]{lyon-thesis}.

\subsection{Problem decomposition}
In order to effect the mapping of this structure onto actual sentences we
decompose the problem into stages, finding the boundaries of one
syntactic feature at a time. The first step is to find the correct
placement for the boundaries of the subject, then further features
are found in the 3 basic constituents. In the current prototype the
head of the subject is subsequently identified. The
processing at each stage is based on similar concepts, and to explain
the role of the neural networks we shall in this paper discuss the first
step in which the subject is found.

The underlying principle employed  each time is to take a sentence,
or part of a sentence, and generate strings with the  boundary markers
of the syntactic constituent in question placed in all possible positions.
Then a neural net selects the string with the correct placement. This is
the grammatical, ``yes'' string for the sentence. The
model is trained in supervised mode on marked up text to find this
correct placement. The different networks that are examined share the
same input and output routines, and each was  integrated into the same
overall system.

\subsection{Tagging}
The first stage in both the training and testing process is to map an 
indefinite number of words onto a limited number of part-of-speech tags.
An automatic tagger 
allocates one or more part-of-speech tags to the words to be processed.
Many words, typically perhaps 25\% to 30\%, have more than one tag. 
The CLAWS automatic
tagger \cite{garside} provided a set of candidate tags for each word, but 
the probabilistic disambiguation modules were not used: disambiguating 
the tags is a sub-task for the neural processer.
The CLAWS tagset was mapped onto a customised tagset of 19, 
used for the work described here.  Further information
on tagset development is in \cite[chapter 4]{lyon-thesis}, and, briefly,
in \cite{lyon6}. 

\subsection{Hypertags as boundary markers}
\label{tag_nums}
As well as part of speech tags we also introduce the syntactic markers, 
virtual tags, which at this stage of
 the process will demarcate the
subject boundary. These {\em hypertags} represent the  opening  
~~{\bf `['}~~and  closing ~~{\bf `]'}~~ of the subject.  The 
hypertags  have relationships with their
neighbours in the same way that ordinary tags do: some combinations are 
likely, some are unlikely. The purpose of the parser is to find the correct
location of the hypertags.

With a tagset of 
19 parts-of-speech, a start symbol and 2 hypertags we have 22 tags in all. 
Thus, there are potentially
$22^2 + 22^3 = 11132$ pairs and triples. In practice only a small proportion 
of tuples are actually realised - see Tables~\ref{newts}~and~\ref{newts2} .
At other stages of the parsing process
larger tagsets are required (see \cite{lyon-thesis}).

\subsection{Rule based pruning: the Prohibition Table}
\label{prohibs}
Strings can potentially be generated with
the hypertags in all
possible positions, in all possible sequences of ambiguous tags.
However, this process
would produce an unmanageable amount of data, so it is pruned by rule based
methods
integrated into the generation process. Applying local and semi-local
constraints the generation of any string is zapped if a prohibited
feature is produced. For fuller details see \cite{lyon-thesis} or
\cite{lyon6}. An example of a local  prohibition  is that the adjacent pair
{\bf (verb,~verb)} is not allowed. Of course {\bf (auxiliary~verb,~verb)} is
permissible, as is {\bf (verb,~~]~~,~verb)}.
These rules are similar to those in a constraint grammar, 
but are not expected to be comprehensive. 
 There are also arbitrary length restrictions on the
sentence constituents: currently, the maximum length of the pre-subject is 15
words, of the subject 12 words.

\subsection{Neural processing}
This pruning operation is powerful and effective, but it still leaves a set
of candidate strings for each sentence - typically between 1 and 25
for the technical manuals. Around 25\% of sentences are left with a single
string, but the rest can only be parsed using the neural selector. This 
averages at about 3 for the technical manuals, more for sentences from 
other domains. In training we manually identify the string with the correct
placement of hypertags, and the correctly disambiguated part-of-speech tags.
In testing mode, the correct string is selected automatically. 

\renewcommand{\baselinestretch}{1}
\normalsize
\clearpage

\subsection{Coding the input}
\label{input-code}
As an example of input coding  consider a short sentence:
\begin{verbatim}
      All papers published in this journal are protected by copyright.  ....(4)

(A) Map each word onto 1 or more tags
            all      predeterminer
         papers      noun  or  verb
      published      past-part-verb
             in      preposition or adverb
           this      pronomial determiner
        journal      noun
            are      auxiliary-verb
      protected      past-part-verb
             by      preposition
      copyright      noun
              .      endpoint

(B) Generate strings with possible placement of subject boundary markers,
    and possible tag allocations (pruned).
string no. 1
strt   [   pred  ]   verb  pastp  prep  prod  noun  aux  pastp  prep  noun  end

.................

string no. 4
strt   [   pred  noun  ]   pastp  adv  prod   noun  aux  pastp  prep  noun  end

string no. 5
strt   [   pred  noun  pastp  ]   adv  prod   noun  aux  pastp  prep  noun  end

string no. 6 *** target ***
strt   [   pred  noun  pastp  prep  prod  noun  ]  aux   pastp  prep  noun  end

string no. 7
strt   [   pred   noun  pastp  adv  prod  noun  ]  aux   pastp  prep  noun  end

(C) Transform strings into sets of tuples

string no. 1
(strt, [ )   ( [, pred )   ( pred, ] ) .......................(noun, end)
(strt, [, pred)   ([, pred, ])   (pred, ], verb)............. (prep, noun, end)

and similarly for other strings
\end{verbatim}

(D) The elements of the binary input vector represent all tuples, initialized
to 0. If a tuple is present in a string the element that represents it 
is changed from 0 to 1. 

\normalsize
\clearpage

\subsection{Characteristics of the data}
\label{charistics}

The domain for which our parser was developed was text from technical manuals
from Perkins Engines Ltd. They were written with the explicit aim of being 
clear and straight forward \cite{perkins}. Using this text as a basis we 
augmented it slightly to develop the prototype on which users try their own 
text. Declarative sentences were 
taken unaltered from the manuals for processing: imperative sentences, 
titles, captions for figures were omitted. 2\% of declarative sentences were
omitted, as they fell outside the current bounds (e.g. the subject had more 
than 12 words). A corpus of 351 sentences was
produced: see Table~\ref{stats}.

\renewcommand{\baselinestretch}{1}
\normalsize
\begin{table}[hbt]
\begin{center}
\begin{tabular}{|l|c|}
\hline
Number of sentences & 351 \\ \hline
Average length & 17.98~words \\ \hline
No. of subordinate clauses: & \\
~~~In pre-subject & 65 \\
~~~In subject & 19 \\
~~~In predicate & 136 \\ \hline
Co-ordinated clauses & 50 \\ \hline
\end{tabular}
\end{center}
\caption{Corpus statistics. Punctuation marks are counted as words, 
formulae as 1 word. \label{stats}}
\end{table}

\normalsize
This corpus (Tr-all) was divided up 4 ways (Tr 1 to Tr 4) so that nets 
could be trained on part 
of the corpus and tested on the rest, as shown in Table~\ref{descrip}.
In order to find the placement of the subject boundary markers we do not
need to analyse the predicate fully, so the part of the sentence being
processed is dynamically truncated 3 words beyond the end of any
postulated closing hypertag. The pairs and triples
generated represent part of the sentence only.

\renewcommand{\baselinestretch}{1}
\normalsize

\begin{table}
\begin{center}
\begin{tabular}{|c|c|c||c|c|c||c|}
\hline
 Training & number of & number of & Test& number of & number of & Ratio of\\
 set      & sentences & strings& set& sentences & strings & testing/training\\
          &           &        &    &           &         & strings   \\ \hline
Tr-all    & 351       & 1037      &        &           &        & \\
Tr 1     & 309       & 852        & Ts 1   & 42       & 85     & 0.10\\
Tr 2    & 292       & 863       & Ts 2 & 59        & 174    & 0.20\\
Tr 3     & 288       & 843       & Ts 3  & 63        & 194    & 0.23\\
Tr 4     & 284       & 825       & Ts 4  & 67        & 212    & 0.26\\ \hline
\end{tabular}
\end{center}
\caption{Description of training and test sets of data \label{descrip}}
\end{table}

\begin{table}
\begin{center}
\begin{tabular}{|l|c|c||l|c|c|}
\hline
 Training& number of    & number of & Test & number of new & number of new  \\
 set     & pairs in     & pairs in   & set & pairs in & pairs in\\
         &`yes' strings & `no' strings & & `yes' strings& `no' strings\\ \hline
 ~Tr-all    & 162          & 213       &      &            &           \\
 ~Tr 1     & 161          & 211        &~Ts 1  & 1 (1\%)    & 2 (1\%)     \\
 ~Tr 2    & 160          & 210         &~Ts 2 & 2 (1\%)   & 3 (1\%)      \\
 ~Tr 3     & 156          & 210        &~Ts 3 & 6 (4\%)    & 3 (1\%)      \\
~Tr 4     & 149          & 193         &~Ts 4 & 13 (9\%)  & 20 (10\%) \\ \hline
\end{tabular} 
\end{center}
\caption{Part-of-speech pairs in training and testing sets.`Yes' indicates correct strings, `no' incorrect ones \label{newts}}
\end{table}

\begin{table}
\begin{center}
\begin{tabular}{|l|c|c||l|c|c|}
\hline
 Training & number of    & number of & Test & number of new & number of new  \\
 set      & triples in   & triples in & set  & triples in & triples in\\
          &`yes' strings & `no' strings &      & `yes' strings& `no' strings\\  \hline
~Tr-all    & 406          & 727          &      &          &         \\
~Tr 1     & 400          & 713          & ~Ts 1  &  6 (2\%)  &  14 (2\%)  \\
~Tr 2     &  383         & 686          & ~Ts 2&  23 (6\%)  & 41 (6\%)   \\
~Tr 3     & 361          & 642          & ~Ts 3 &  45 (12\%) &  85 (13\%)    \\
~Tr 4      & 364          & 632        & ~Ts 4 &  42 (12\%) & 95 (15\%)  \\ \hline
\end{tabular}
\end{center}
\caption{Part-of-speech triples in training and testing data sets 
\label{newts2}}
\end {table}

\begin{table}
\begin{center}
\begin{tabular}{|l|c|}
\hline
Total number of pairs in `yes' strings & 4142\\
Total number of pairs in `no' strings  & 8652 \\
Total number of triples in `yes' strings & 3736 \\
Total number of triples in `no' strings & 8021 \\ \hline
\end{tabular}
\caption{ Total number of tuples in Tr-all, including repetitions
\label{newts3}}
\end{center}
\end{table}

\normalsize
\subsection{Data distribution}

\begin{figure}[hbt]
\begin{center}
\strut\psfig{figure=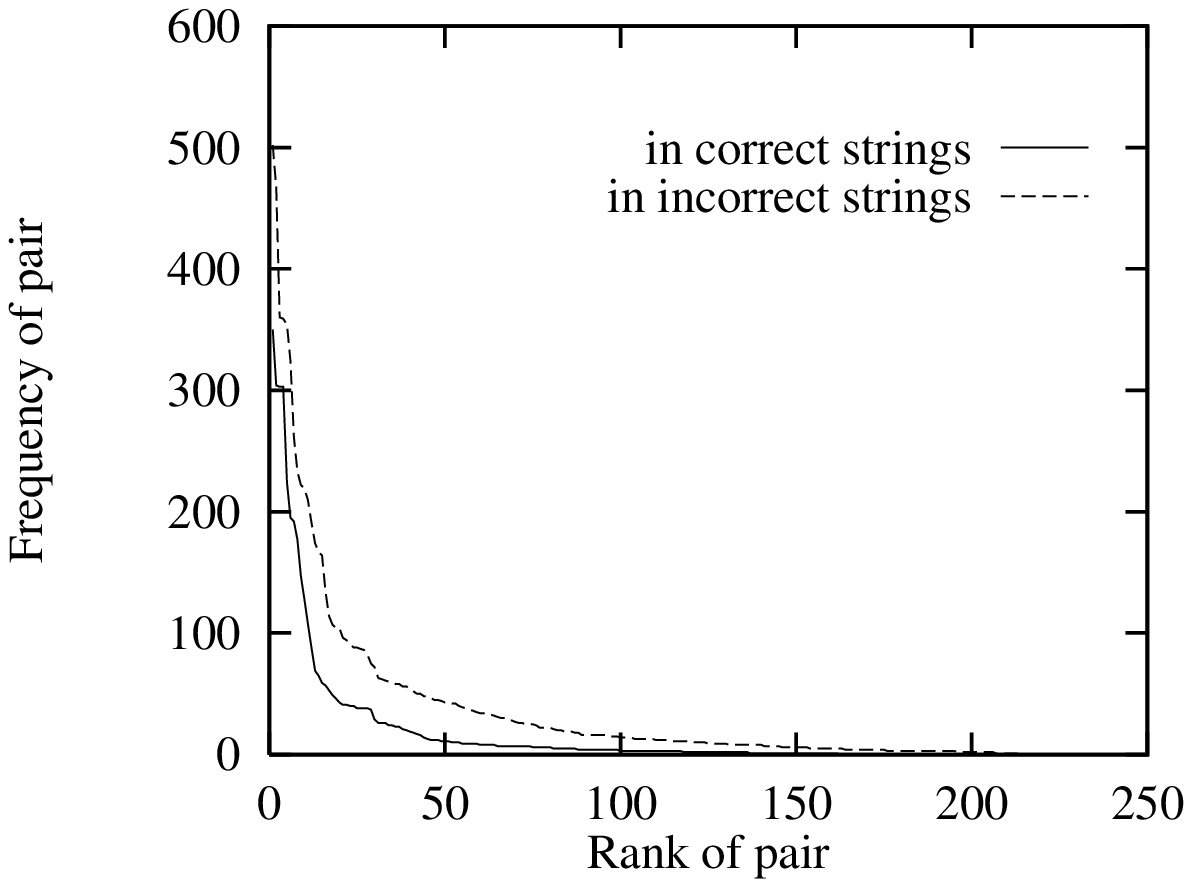,width=4in}
\end{center}
\caption{Data from 351 sentences in technical manuals. Pairs are ranked by
frequency of occurrence in correct and 
incorrect strings. Relationship between rank and frequency shown.\label{graph_a}}
\begin{center}
\strut\psfig{figure=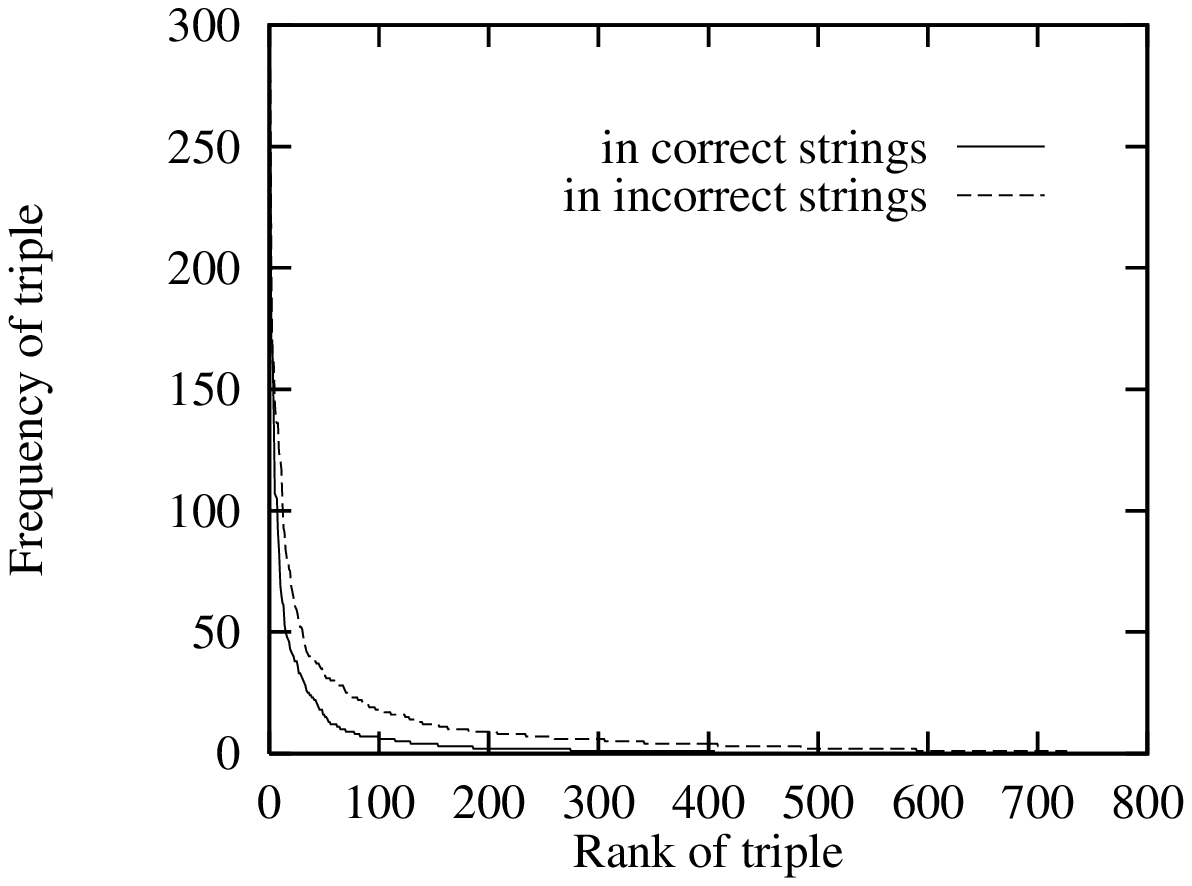,width=4in}
\end{center}
\caption{Relationship between rank and frequency of triples on the same data.\label{graph_b}}
\end{figure}

Statistics on the data generated by the Perkins corpus are given in 
Tables~\ref{newts}, \ref{newts2}~and~\ref{newts3}.
A significant number of tuples occur in the test set,
but have not occurred in the training set, since, as 
Figures~\ref{graph_a}~and~\ref{graph_b} show, the distribution of 
data has a zipfian character. 

\clearpage
\clearpage

\subsection{Interpreting the output } 
\label{output}

For training, the set of strings generated by the training text is taken as a
whole. Each
string is given a ``grammaticality measure'' {\bf $\Gamma$}, where 
{\bf $\Gamma$} is  positive for a correct string, negative for an incorrect
one. Details follow in Section~\ref{single}.
In testing mode we
consider separately the set of strings generated for each sentence,  
and the string with the highest {\bf $\Gamma$} measure is taken as the 
correct one for that sentence. In the example in Section~\ref{input-code}
string  6 is the correct one.

However, on test data, there are 4 metrics for correctness that can be useful 
in different practical situations.
The measure ``correct-a" requires only that the hypertags are correctly 
placed: string 7 as well as string 6 is correct-a. ``correct-b" requires also 
that  all words within the subject
are correctly tagged, ``correct-c" that all words 
within the part of the sentence being processed are correctly tagged. 
The final measure
 ``correct-d" records the proportion of strings that are
in the right class. It can happen that the highest scoring string may have 
a negative {\bf $\Gamma$}. Conversely, some incorrect strings can have a 
positive {\bf $\Gamma$} without having the highest score. 
For practical purposes the measures ``correct-a'', ``-b'', and ``-c''
will be the significant ones. But in analysing the performance of the 
networks we will be interested in ``correct-d'', the extent to which the net
can generalize and correctly classify strings generated by the test data.

\subsection*{Metrics of other systems}
Note that these measures relate to a string, not to individual elements
of the string. This contrasts with some natural language processing systems,
in which the measure of correctness relates to each word. For instance,
automated word tagging systems typically measure success by the proportion
of {\em words} correctly tagged. The best stochastic taggers typically 
quote success rates of $95\%$ to $97\%$ correct. If sentences 
are very approximately 20 words long on average, this can mean that there
is an  error in many  sentences.
\section{Using single layer networks}
\label{lin-sep}

\subsection{Conversion to linearly separable forms}
\label{Phi function}
It is always theoretically possible to solve supervised learning
problems with a
single layer, feed forward network, providing the input data is enhanced
in an appropriate way. A good explanation is given by 
Pao~\cite[chapter 8]{pao2}. 
Whether this is desirable in any particular case 
must be investigated.   The enhancement can map the input data onto a space, 
usually of higher
dimensionality,  where it will be linearly separable. Widrow's valuable 1990
paper on ``Perceptron, Madaline, and Back Propagation"
\cite[page 1420]{widrow1} explores these approaches
``which offer great simplicity and beauty".  

\begin{figure}[hbt]
\begin{center}
\strut\psfig{figure=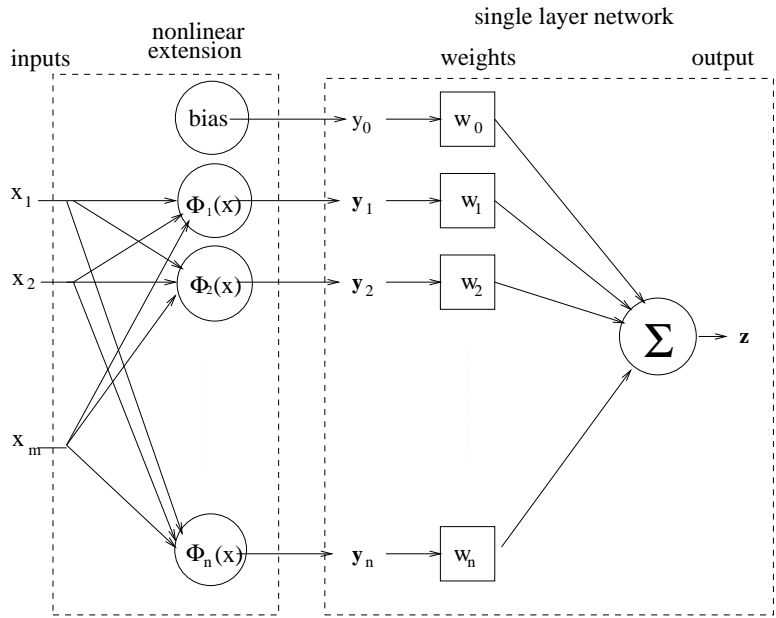,width=4in}
\end{center}
\caption{The Generalized Single Layer Network, GSLN, with 1 output
\label{gsln2}}
\end{figure}

Figure~\ref{gsln2} illustrates the form of the Generalized Single Layer
Network (GSLN). This figure is derived from Holden and Rayner \cite{holden}.
A non-linear transformation $\Phi$ on inputs $\{x\}$  converts them to 
elements $\{y\}$, which a
single layer net will then process to produce an output 
$z$ (temporarily assuming 1 output). 
The  $\Phi$ functions, or  basis functions,
can take various forms.
They can be applied to each input node separately, or, as indicated in the 
figure, they can
model the higher order effects of correlation. In our processor $\Phi$ is 
an ordered `AND', described in the following section. A similar function is  
used  in the grammatical inference work of Giles et
al. \cite{giles1}; it is also used in DNA sequence analysis 
\cite{lapedes}. The $\Phi$ function can  be arithmetic: for 
instance, for  polynomial discriminant functions   the elements of the 
input vectors are combined as products \cite[page 135]{duda}. 
Successful uses of this approach include the discrimination of different
vowel sounds~\cite{holden2} and the automated
interpretation of telephone company data in tabular form~\cite{chhabra}.

Radial basis function (RBF) networks also come into the class of GSLNs. 
In their two stage training procedure  the parameters governing the basis 
functions are determined first. Then a single layer net is
used for the second stage of processing. Examples include Tarassenko's
work on the analysis of ECG signals~\cite{tarassenko} and a solution to an 
inverse scattering problem to determine particle properties~\cite{wang}.  

An important characteristic of the GSLN is
that processing at different layers is de-coupled. The first stage of training
is unsupervised: the $\Phi$ functions are applied without recourse to
desired results. In the second stage of training a supervised method is
used, in which weight adjustments on links are related to target outputs. 

One perspective on the GSLN  is given by 
Bishop~\cite[e.g. page 89]{bishop}, who
characterises this type of system as a special case of the more general
multi-layer network.  Whereas in the general case the basis functions at the
first layer are modified during the training process, in this type of 
system the basis functions are fixed independently. 
Widrow~\cite[page 1420]{widrow1} puts it the other way
round: ``one can view multi-layer networks as single layer networks with
trainable preprocessors...''. 
\clearpage
\subsection{The conversion function used in the parser}

We use this approach in the parser by converting a sequence of tags 
to a higher order set of adjacent pairs and triples.
The example in section \ref{input-code}, stage C shows
how the input elements are constructed. Thus, one of the elements derived 
from string 1 is {\bf (~[,~predeterminer,~verb~)}. This of course is not 
the same as {\bf (~predeterminer,~[,~verb~)}.
The same tag can be repeated within a tuple: {\em Computer Science Department}
maps onto {\bf (~noun,~noun,~noun~)}.

This can be related to Figure \ref{gsln2}. Let each $x_i$ for $i = 1$ to $m$
represent a tag. The $\Phi$ functions map these onto $m^2$ pairs and  $m^3$ 
triples, so  $n = m^2 + m^3$.  
If $s$ is a sequence of $l$ tags, it is transformed into a set $S$ of 
higher order elements by $\Phi_p$ for pairs and $\Phi_t$ for triples.

\renewcommand{\baselinestretch}{1}
\normalsize

 \[ s = x_1 \ldots x_{i}, x_{i+1}, x_{i+2} \ldots x_l\]

\[ \Phi_p(x_i) = (x_i,x_{i+1})  \qquad\mbox{ for } i=1 \mbox{ to } i=l-1 \]

\[ \Phi_t(x_i) = (x_{i},x_{i+1},x_{i+2})\qquad\mbox{ for } i=1 
\mbox{ to } i=l-2 \]

\[ S = \{ \Phi_p(x_i) \} \cup \{ \Phi_t(x_i) \} \]

\normalsize 
For some of our investigations either pairs or triples were used, rather 
than both.

The $\Phi$ function represents an ordered `AND': the higher order elements  
preserve sequential order.
This function was derived using heuristic methods, but the approach was
supported by an objective analysis of the proposed representation.
 We aimed to capture {\em some} of the implicit 
information in the data, model invariances, represent structure. 
As described in Section~\ref{tuples}, the choice of the tupling 
pre-processing function is supported by information theoretic analysis. 
It captures local, though not distant, dependencies. Using this representation
we address simultaneously the issues of 
converting data to a linearly separable form, modelling its sequential 
character and capturing some of its structure.

An approach similar
to our own has been used to develop neural processors for an analysis
of DNA sequences \cite[page 166]{lapedes}. Initially a multi-layer network 
was used for one task, but an analysis of its
operation led to the adoption of an improved representation with a simpler 
network. The input representing bases 
was converted to codons (tuples of adjacent bases), and then processed with a 
Perceptron.

\subsection{The practical approach}
\label{minsky}
Minsky and Papert acknowledged that single layer networks could classify
linearly inseparable data if it was transformed to a sufficiently high
order \cite[page 56]{minsky}, but claimed this would be impractical. 
They illustrated the point with the example of a parity tester. 
However, this example is the extreme case,
where {\em any} change of a single input element will lead to a change in the
output class. If there are $n$ inputs, then it is necessary to tuple each
element together to O($n$). The consequent  explosion of input data 
would make the method unusable for all but the smallest data 
sets.

However, in practice, real world data may be different.
Shavlik et al \cite{shavlik} compare
single and multi-layer nets on 6 well known problems, and conclude
``Regardless
of the reason, data for many `real' problems seems to consist of
linearly separable categories..... Using a Perceptron as an initial test
system is probably a good idea." 
This empirical approach is advocated here. Tests for linear separability
and related problems are computationally heavy \cite{sklansky,lin}, so we 
 tried single layer
networks to see whether the higher order data we use is in practice
linearly separable, or nearly so.

Taking data items as pairs typically produces training
sets of which about 97\% can be learnt by a single layer network; taking 
triples raises learnability to about 99\%. 
Thus our data is almost linearly separable.

\subsection{Linear discriminants - neural and Bayesian methods}

Having established empirically that after transformation we have a linear
problem, there are a number of different methods of linear discriminant
analysis that could be used. Our single layer networks are convenient
tools.      

We also ran our data through a Bayesian classifier, based on the 
model described by Duda and Hart \cite[page 32]{duda}. Results were about 
5\% less good on test data than those from Hodyne. Though the parsing
problem is decomposed so that good estimates can usually be made of
prior probabilities, estimating class conditional probabilities needs further 
investigation. (If $n$ possible parses
are generated and 1 is correct, then the prior is $1/n$ ). 
Frequency counts extracted from the training data cannot be used as they
stand as probability estimates. The zipfian distribution of data can distort 
the probabilities, even when very large quantities are used, so that rare
events are given too much significance. Moreover, further information on
zero frequency items, though limited, can
be extracted using an appropriate technique, as Dunning shows \cite{dunning}.
There are a number of methods of estimating probabilities on the basis
of partial information which need investigation \cite[page 55]{bell}.

These issues can be avoided by using neural discriminators.

\subsection{Training set size}
There is a relationship between training set size and linear separability.
Cover's classical work addressed the probability that a set of {\em
random}, real valued vectors with random binary desired responses are 
linearly separable \cite{cover,widrow1}. Using his terminology and taking 
the term ``pattern'' to mean a training example, the critical factor is 
the ratio of number of patterns, $\Pi$, to
number of elements in each pattern, $n$. While $\Pi/n < 1$, the
probability $P_{separable} = 1.0$.
If $\Pi/n = 1$  then $P_{separable} = 0.5$. As $\Pi/n$ increases,
$\Pi_{separable}$ quickly declines. 

These observations are given as background information to indicate that
training set size should be considered, but they do not apply 
in our case as they stand. First, our data is not random. 
Secondly, a necessary condition that the vectors are in ``general position'',
normally satisfied by real valued 
vectors, may not hold for binary vectors \cite[page 97]{hertz}.

The number of training examples, $\Pi$, is a  factor in determining
generalization capability (see, for example, \cite{baum,holden}). The 
probability that an error is within a certain bound increases with
the number of training examples. Decreasing 
$\Pi$ to convert data to a linearly separable form would be profitless.

\begin{figure}[hbt]
\begin{center}
\strut\psfig{figure=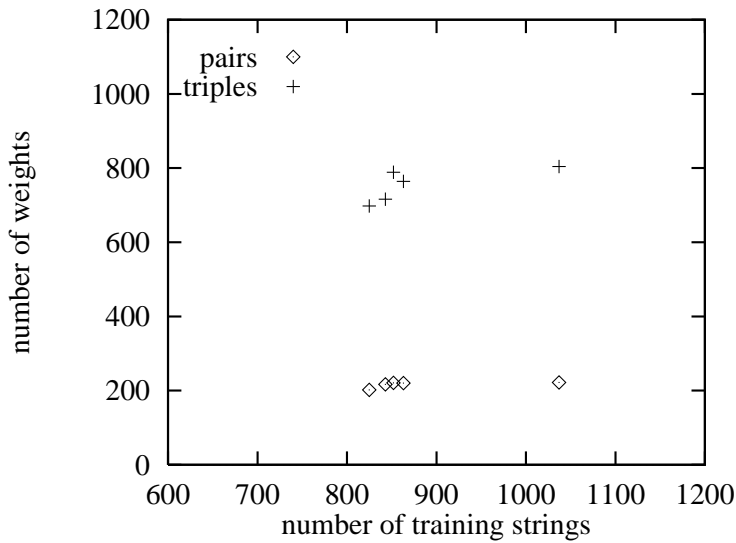,width=4in}
\end{center}
\caption{Relationship between number of training examples and number of
weights, for the Perceptron and LMS nets with one output. \label{tr_size2}}
\end{figure}

The ratio of  training examples to weights in our data is shown in 
Figure~\ref{tr_size2}.
Note that the corpus used for this preliminary working prototype is small 
compared to other corpora, and future work will use much larger ones, which
could affect  this ratio. 

\section{Three single layer networks}
\label{single}
\subsection{Architecture}

\begin{figure}[hbt]
\begin{center}
\strut\psfig{figure=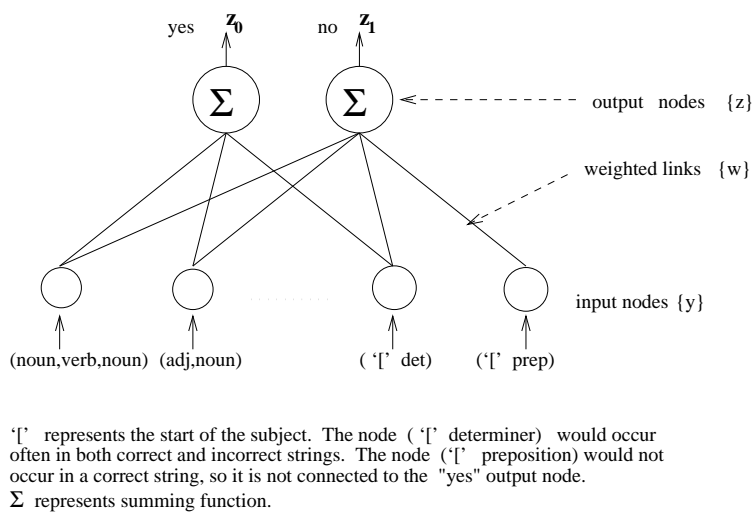,width=4in}
\end{center}
\caption{The Hodyne network. \label{hodyne}}
\end{figure}

Refer again to Figure~\ref{gsln2}, illustrating the GSLN. In this work 
we compare 3 networks, which all use 
the same $\Phi$ functions,  described in 
Section~\ref{Phi function}. We now compare 
methods of processing at the second stage, that is the performance of 3
different single layer classifiers. The Perceptron and LMS net
can be characterised as examples of the GSLN in Figure~\ref{gsln2}.
The Hodyne model differs in having 2 outputs, not symmetrically
connected, as in Figure~\ref{hodyne}.

\subsection{Methods of adjusting connection weights during training}
\label{methods}
When single layer networks are used, we do not have the classic problem
of ``credit assignment'' associated with multi-layer networks: the
input neurons responsible for incorrect output can be identified, and 
the weights on their links adjusted. There is a choice of methods for
updating weights. We do not 
{\em have} to use differentiable activation functions, as in the multi-layer
Perceptron. These methods can therefore be
divided into two broad categories. First there are ``direct update'' methods,
used in the traditional Perceptron and Hodyne-type nets, where a weight update
is only invoked if a training vector falls into the wrong class. 
This approach is related to the ideas behind reinforcement learning, but
there is no positive reinforcement. If the 
classification is correct weights are left alone. If the classification
is incorrect then the
weights are incremented or decremented. No error measure is needed:
the weight update is a function either of the input
vector (Perceptron), or of the existing  weights (Hodyne).

Secondly, there are ``error minimization'' approaches, which can also
be used in multi-layer nets. 
An error measure, based on the difference between 
a target value and the actual value of the output, is used.
This is frequently, as in standard back propagation, a process
based on minimizing the  mean square  error, to reach the LMS error \cite{pdp}.
We have used a modified error minimization method (Section~\ref{LMS}).

\subsection{The Perceptron}
The Perceptron and LMS models are both fully connected nets with a single 
output. Details of the well known Perceptron training algorithm,
 and of the parameters used, are in \cite{Lyon10}. 
 The output represents $\Gamma$, the grammaticality measure of the 
input string. In training, a grammatical string must produce a positive
output, an ungrammatical string a negative output. The wrong output
triggers a weight adjustment on all links contributing to the result. This is
 a function of the normalized input values, scaled by the learning rate.

To speed the training process, a method of ``guided initialization'' sets 
initial random weight 
within bounds determined by the expected output. To implement this,
see whether a new, previously unseen  input element belongs 
to a ``yes'' or ``no'' string, 
corresponding to desired positive or negative outputs. Then set a random value
between $0.0$  and $0.3$ for ``yes'', between $0.0$  and $-0.3$ for ``no''
\cite{lyon-thesis}. When training is finished, weights on links from unvisited
input elements are set to 0.0.

Recall that in testing mode we
consider the set of strings generated for each sentence.
and the string with the highest {\bf $\Gamma$} measure is taken as the
correct one for that sentence. 

\subsection{Hodyne}

This network, shown in Figure~\ref{hodyne}, is derived from the model 
introduced by Wyard and Nightingale
\cite{wyard}. The 2 outputs $z_0$ and $z_1$ represent grammatical and
ungrammatical, ``yes'' and ``no'', results. In training, a grammatical 
string must produce $z_0 > z_1$, and vice-versa, else a weight adjustment
is invoked. In testing mode, as for the Perceptron, 
the strings generated by each sentence are processed, and
the string with the highest $\Gamma$ score for that sentence is the winner.
For this network the grammaticality measure $\Gamma$ is
$z_0 - z_1$. 
Since it is not widely known a summary of the training method 
follows. More implementation details can  be found in \cite{lyon-thesis}.

\clearpage
\noindent
Notation

\renewcommand{\baselinestretch}{1}
\small

\noindent
Let each input vector have $n$ elements $y_i$.
Let $w(t)_{i,j}$ be the weight from the 
{\em i}th input node to the {\em j}th output node at time $t$. 
Let $u(t)_{i,j}$ be the update factor.

\noindent
$\delta = -1$ or $\delta = +1$ 
indicates whether weights should be decremented or incremented. 
\begin{quotation}
\sf
\small\sf
\begin{tabbing}
\hspace*{1cm}\=\hspace{1cm} \=hspace{1cm}\=hspace{1cm} \kill\\
Mark all links disabled.\\
Initially, percentage of strings correctly 
classified = 0.0\\
REPEAT \\
\quad from START1 to END1 until \%  strings correctly classified
exceeds chosen threshold:\\
START1\\
\> REPEAT from START2 to END2 for each string\\
\>START2 \\
\> \>Present input, a binary vector,  $y_1, y_2, .... y_n$\\
\> \> Present  desired output, $z_0 > z_1$ or {\em vice versa}\\
\> \> For any $y > 0$ enable link to desired result if it is disabled\\
\> \> Initialize weight on any new link to 1.0\\
\> \> Calculate outputs $z_0$ and $z_1$\\
\> \> \> \( z_k = \sum_{i=1}^{i=n} w_{i,k} \ast y_i \)\\
\> \> If actual result = desired result\\
\> \> \> Count string correct, leave weights alone \\
\> \> Else adjust weights on current active links:\\
\>\> \> if $z_0 < z_1$  then $\delta = +1$ on links to $z_0$, $\delta = -1$
on links to $z_1$ \\
\> \> \> and {\em vice versa}\\
\end{tabbing}
 \[ w(t+1)_{i,j} = \left[1 +\frac{\delta\ast w(t)_{i,j}}{1+(\delta \ast w(t)_{i,j})^4}\right]w(t)_{i,j}\]
\begin{tabbing}
\hspace*{1cm}\=\hspace{1cm} \=hspace{1cm}\=hspace{1cm} \kill\\
\>END2\\
\>Calculate \% strings correctly classified. If greater than threshold, terminate \\
END1
\end{tabbing}
\end{quotation}
\vspace{0.5cm}
\normalsize
\noindent
  
 For the Hodyne type net the update factor $u$ is a function
of the current weights; as the weights increase it is 
asymptotic to $0$, as they decrease it becomes equal to $0$. 
\[ u(t)_{i,j} =  \frac{\pm w(t)_{i,j}^2}{1 \pm w(t)_{i,j}^4}  \]

\begin{figure}[hbt]
\begin{center}
\strut\psfig{figure=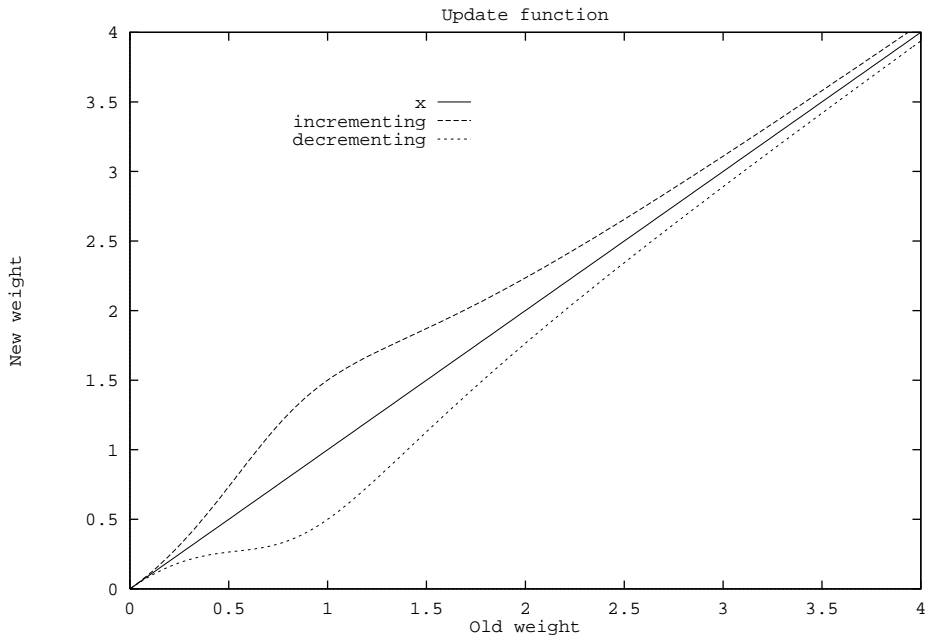,angle=270,width=5.0in}
\end{center}
\caption{Relationship between old and new weights for the Hodyne net
\label{trad-hod}}
\end{figure}

This function satisfies the requirement that the
weights increase monotonically and saturate (see Figure~\ref{trad-hod}). 
We use the original Hodyne function with a comparatively  low computational 
load. Note that in contrast to the
Perceptron, where the learning rate is set at compile time, the effective
learning rate in this method varies dynamically. The greatest changes occur
when weights are near their initial values of 1.0, as they get larger or 
smaller the weight change decreases. 

\clearpage

\subsection*{Hodyne's pattern of connectivity and the Prohibition Table}
\label{linkage}

Note that during training elements of a new input vector may be linked to
both, either or neither output node. This represents the fact that a tuple can
appear (i) in both a correct and incorrect string, (ii) in either or
(iii) in neither. Tables~\ref{newts}~and~\ref{newts2} gives some 
information on the distribution of elements in training and testing sets.
The data is asymmetric: any node that appears in a 
grammatical string can also appear in an ungrammatical one, but the reverse
is not true. 
When training is completed links from unused inputs are enabled
and their weights set to 0.0

Any single entry in the preliminary Prohibition Table (Section~\ref{prohibs}) 
can be omitted, and 
the pair or triple be included in the neural processing task. In this case a 
tuple that cannot occur in a grammatical string will, for Hodyne, only be 
connected to the ``no'' output
node. Conversly, if we examine the linkage of the Hodyne net, those tuples
that are only connected to the ``no'' output are candidates for inclusion in
a constraint based rule. Of course there is a
chance that a rare grammatical occurrence may show up as the size of the 
training set increases.
%
%

\subsection{LMS network}
\label{LMS}
The LMS model is 
based on the traditional method, described in ``Parallel
Distributed Processing'' \cite[page 322]{pdp}. A bipolar activation function 
is used, and outputs are in the range $-1$
to $+1$. As with the Perceptron, the output represents the $\Gamma$ measure
for the string being processed. Gradient descent is used to reduce the error
between desired and actual outputs.
 
It has been known for
many years in the numerical optimization field that the gradient descent 
technique is a
poor, slow method \cite{dixon}. This is now also accepted wisdom in the
neural network community \cite{bishop}. Other training methods, such
as conjugate gradients, are usually preferable. For this experiment, however,
the traditional method has been used, but some variations to speed
up training and improve performance have been incorporated.
Brady et al.~\cite{brady} described some anomalies that can arise
with the traditional LMS model. As a remedy Sontag's technique for 
interpreting the error measure is included \cite{sontag}.
This means that an error is only recorded if a vector falls into the wrong
class. The target output is a threshold, and if this threshold is 
passed the vector is 
considered correctly classified. This contrasts with the original  LMS
method, in which  an error is recorded if the target is either undershot or
overshot. 

\section{Performance}
\label{results}
\subsection{Training times}
Since we have been developing a prototype the training threshold was taken so 
that training was fast - less than 10 seconds for the Perceptron, less than
20 seconds for the Hodyne network. Subject to this constraint the percentage
of strings that could be trained ranged from $96.5\%$ to $99.0\%$. 
The Perceptron was fastest. For the LMS net training times were between 1 and
2 orders of magnitude greater, but the inefficient gradient descent 
method was used. 

\subsection{Ability to generalize}
Results are given in Tables~\ref{results1} to  \ref{results3}.
This system has been developed to produce a winning string for each sentence,
and performance can be assessed on different measures of correctness,
as described in Section~\ref{output}. For the purpose of investigating
the function of the networks we take the strictest measure, {\em correct-d}
in the tables, requiring that strings should be classified correctly.
However, we can interpret the results so that in practice we get  up to
100\% correct for our practical application, since a winning string
may have a negative $\Gamma$ measure. Thus the practical measure of
correctness can be higher than the percentage of correctly classified strings.

\renewcommand{\baselinestretch}{1}
\normalsize

\begin{table}
\begin{center}
\begin{tabular}{|c|c|c|c|c||c|c|c|c|}
\hline
Training&Test&Pairs&Triples&Training&correct-a&correct-b&correct-c&correct-d \\
set     & set &used& used  &threshold&  \%    & \%      &  \%     &      \% \\ \hline 
Tr 1     & Ts 1 & Y  &  Y    & 95.0    &100     & 97.6    &  95.2   & 92.9  \\
        &     & Y  &  Y    & 99.0    &100     & 97.6    &  95.2   & 89.4 \\
        &     & Y  &       & 97.0    &100     & 97.6    &  92.9   & 85.8 \\
        &     &    &  Y    & 98.5    & 100    & 97.6    &  97.6   & 91.8 \\ \hline
Tr 2   &Ts 2& Y  &  Y    & 99.0    &96.6    & 96.6    &  91.5   & 88.5 \\
	&     & Y  &       & 98.0    &93.2    & 93.2    &  88.1   & 89.1 \\
	&     &    &  Y    & 98.5    &98.3    & 98.3    & 89.8    &  85.6\\ \hline
Tr 3    & Ts 3& Y  &  Y    &99.0     & 98.4   & 98.4    &  95.2   & 80.9 \\
        &     & Y  &       &96.0     & 96.8   & 93.7    &  92.1   & 85.6 \\
        &     &    &  Y    & 99.0    & 96.8   & 95.2    &  90.5   & 78.9 \\ \hline
Tr 4    & Ts 4& Y  &  Y   & 99.0     &92.5    & 92.5    & 74.6    & 77.4 \\
        &     & Y  &      & 97.0     &92.5    & 92.5    & 92.5    & 82.5 \\
        &     &    & Y    & 99.0     &89.6    & 88.1    &  83.6    & 78.3 \\
\hline
\end{tabular}
\end{center}
\caption{Results using Perceptron.
Recall that correct-a means hypertags 
are correctly placed, correct-b that words inside subject are correctly
tagged also, correct-c that all words in part of sentence being processed are
also correctly tagged, correct-d that the string is in the right class 
\label{results1}}
\end{table}
\begin{table}
\begin{center}
\begin{tabular}{|c|c|c|c|c||c|c|c|c|}
\hline
Training&Test&Pairs&Triples&Training&correct-a&correct-b&correct-c&correct-d \\
set     & set &used& used  &threshold&  \%    & \%      &  \%     &      \% \\
\hline
Tr 1     & Ts 1 & Y  &  Y    & 95.0    &100    & 100      & 97.6     & 91.8 \\
        &     & Y  &  Y    & 99.0    &100    & 100      & 100      & 92.9 \\
        &     & Y  &       & 97.0    &100    &  97.6     & 95.2     & 90.6 \\
        &     &    &  Y    & 98.5    &100    &  100      & 100      & 89.4 \\ \hline
Tr 2   & Ts 2& Y &  Y    & 99.0    & 100   & 100       & 94.9     & 91.4 \\
        &     &  Y &       & 98.0    & 100   & 100       & 94.9     & 90.2 \\
        &     &    &  Y    & 98.5    & 98.3  & 98.3      & 94.9     & 96.8 \\ 
\hline
Tr 3    & Ts 3& Y  &  Y    & 99.0    & 100   &  98.4     & 95.2     & 89.2 \\
        &     & Y  &       & 96.0    & 98.4  &  98.4     & 95.2     & 91.2  \\
        &     &    &  Y    & 99.0    & 96.8  &  96.8     & 92.1     & 86.1 \\  \hline
Tr 4	&Ts 4& Y  & Y     & 99.0    & 95.5  & 94.0      & 91.0     & 84.4   \\
        &     & Y  &       & 98.0    & 95.5  & 94.0      & 92.5     & 83.0 \\
	&     &    & Y     & 99.0    & 94.0  & 94.0      & 83.6     & 82.5 \\
\hline
\end{tabular}
\end{center}
\caption{Results for Hodyne net on same training and testing data
as for the Perceptron (Table~\ref{results1}) \label{results2}}
\end{table}
\begin{table}
\begin{center}
\begin{tabular}{|c|c|c|c|c||c|c|}
\hline
Training&Test&Pairs&Triples&Training&correct-c&correct-d \\
set     & set &used& used  &threshold&  \%    & \%       \\ 
\hline
Tr 1     & Ts 1 & Y     &  Y &  99.0    & 97.6     & 92.9 \\
        &     & Y     &    &  97.0    & 90.5     & 88.3 \\
        &     &       &  Y &  99.0    & 97.6     & 90.6 \\
\hline
Tr 2   & Ts 2& Y    & Y  &  99.0    &91.5      & 88.5 \\
        &      & Y    &    &  98.0    &88.1      & 88.5 \\
        &      &      & Y  &  98.5    &93.2      & 85.1 \\
\hline
Tr 3    & Ts 3& Y     & Y  &  99.0    & 93.7     & 85.1 \\
        &     & Y     &    &  96.0    & 95.2     & 87.6 \\
        &     &       & Y  &  99.0    & 90.5     & 80.9 \\
\hline
Tr 4    & Ts 4 & Y    & Y  &  99.0    &92.5      & 82.5\\
        &      & Y    &    &  97.0    & 97.0     & 83.5\\
        &      &      & Y  &  99.0    & 85.1     & 80.2\\
\hline
\end{tabular}
\end{center}
\caption{Results for LMS net on the same data. Compare with 
Tables~\ref{results1}~and~\ref{results2} \label{results3}}
\end{table}

\normalsize

Table~\ref{summary} gives a summary of the results, showing how these vary
with the ratio of test set size to training set size. This would be expected.
If there is insufficient training data performance degrades sharply.

The Hodyne net performed well, and this 
architecture was used for the prototype.
Previous work in this field compared the performance of multi-layer 
Perceptrons to that of single layer models, and found they performed less well.
This is discussed in Section \ref{op-of-net}.

\renewcommand{\baselinestretch}{1}
\normalsize

\begin{table}
\begin{center}
\begin{tabular}{|c||c|c|c||c|}
\hline
Ratio test set $/$ & Perceptron & Hodyne      & LMS            & Hodyne \\
training set &\% test strings& \% test strings&\% test strings &\% hypertags\\
             &   correct     &  correct       & correct        & correct \\ \hline
0.10            &     89.4   &  92.9  & 92.9 & 100\\
0.20            &     88.5   &  91.4  & 88.5 & 100\\
0.23            &     80.9   &  89.2  & 85.1 & 100 \\
0.26            &     77.4   &  84.4  & 82.5 & 95.5 \\ \hline
\end{tabular}
\caption{Summary of results culled from Tables \ref{descrip}, \ref{results1}, \ref{results2} and \ref{results3}, showing performance on 4
different training and test sets. \label{summary}}
\end{center}
\end{table}

\clearpage
\normalsize
\section{Understanding the operation of the network}
\label{op-of-net}
\renewcommand{\baselinestretch}{1}
\normalsize

\subsection{The importance of negative information}
Consider the following unremarkable sentence,  and some of the strings it 
generates:
\begin{verbatim}
            the      determiner
     directions      noun
          given      past-part-verb
          below      preposition  or  adverb
           must      auxiliary verb
             be      auxiliary verb
      carefully      adverb
       followed      past-part-verb
              .      endpoint
string no. 1:
strt    [    det   noun   ]     pastp  adv    aux   aux   adv    pastp  endp

string no. 2:
strt    [    det   noun  pastp   ]    prep    aux   aux   adv    pastp  endp

string no. 3: *** target ***
strt    [    det   noun  pastp  prep   ]     aux   aux   adv    pastp  endp
\end{verbatim}

\normalsize

In the LOB corpus the pair {\bf (preposition, modal-verb)},
which represents the words {\bf (below, must)}\footnote{Modal verbs are 
included in the class of auxiliary verb in this tagset} 
has a frequency of less than $0.01\%$, if it occurs at all~\cite{johansson}. 
So when a sentence like this is processed in testing mode the particular 
construction may well
not have occurred in any  training string. However, in the
candidate strings that are generated wrong placements
should be associated with stronger negative weights {\em somewhere} in 
the string. For example,  string 2 maps onto:
\begin{quote}
* [ The directions given ] below must be carefully followed.
\end{quote}
The proposed subject would not be associated with strong negative 
weights. However, the following pairs and triples include at least  one that 
is strongly negative, such as {\bf( ], preposition)},
an element in the negative strings
generated in the training set. The correct placement, as in string 3, would be 
the least bad, the one with the highest  $\Gamma$ score.

By training on negative as well as positive examples we increase the 
likelihood that in testing mode a previously unseen structure can be correctly
processed. 
In this way the probability of correctly processing rare constructions is
increased.

\subsection{Relationship between frequency of occurrence and weight}

\begin{figure}[hbt]
\begin{center}
\strut\psfig{figure=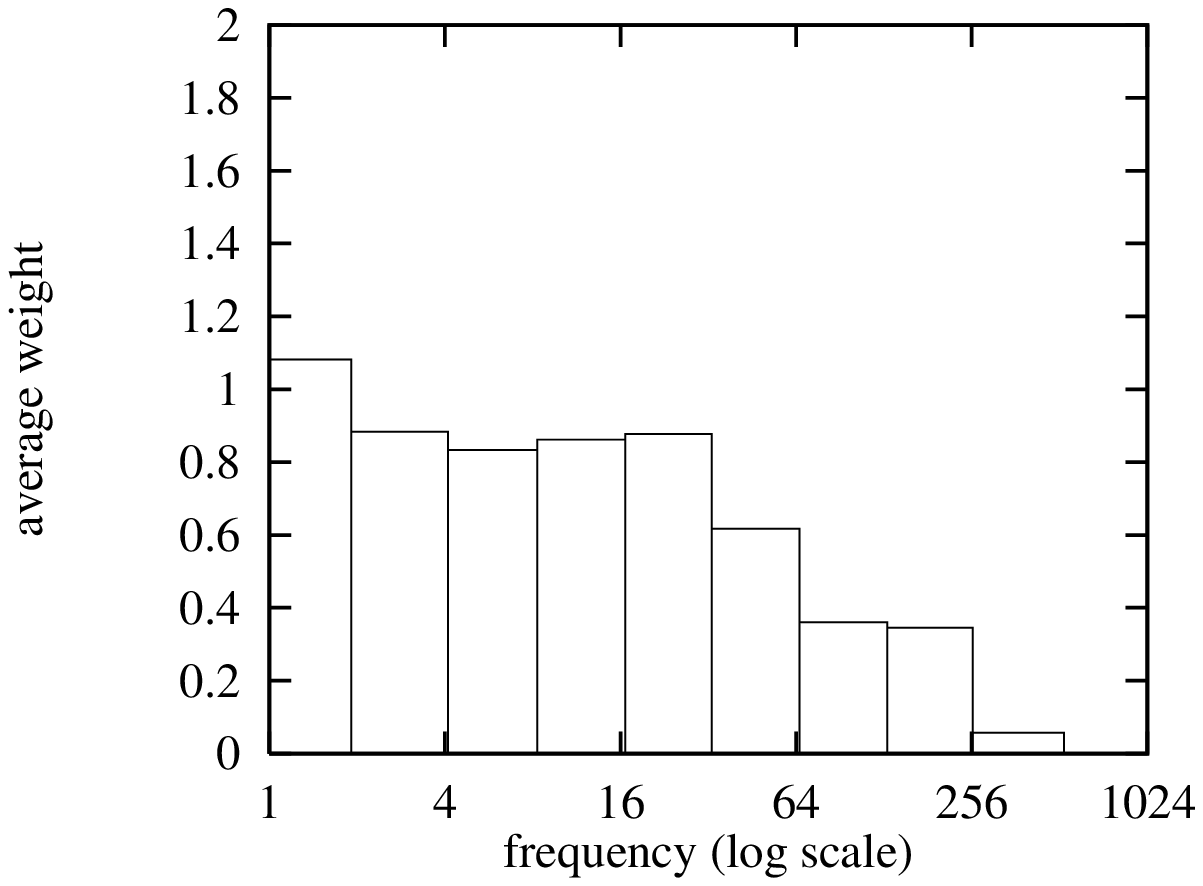,width=3.6in}
\caption{Weights plotted against frequency of input node occurring, for training corpus Tr 1 on Hodyne \label{hod-wts}}
\end{center}
\begin{center}
\strut\psfig{figure=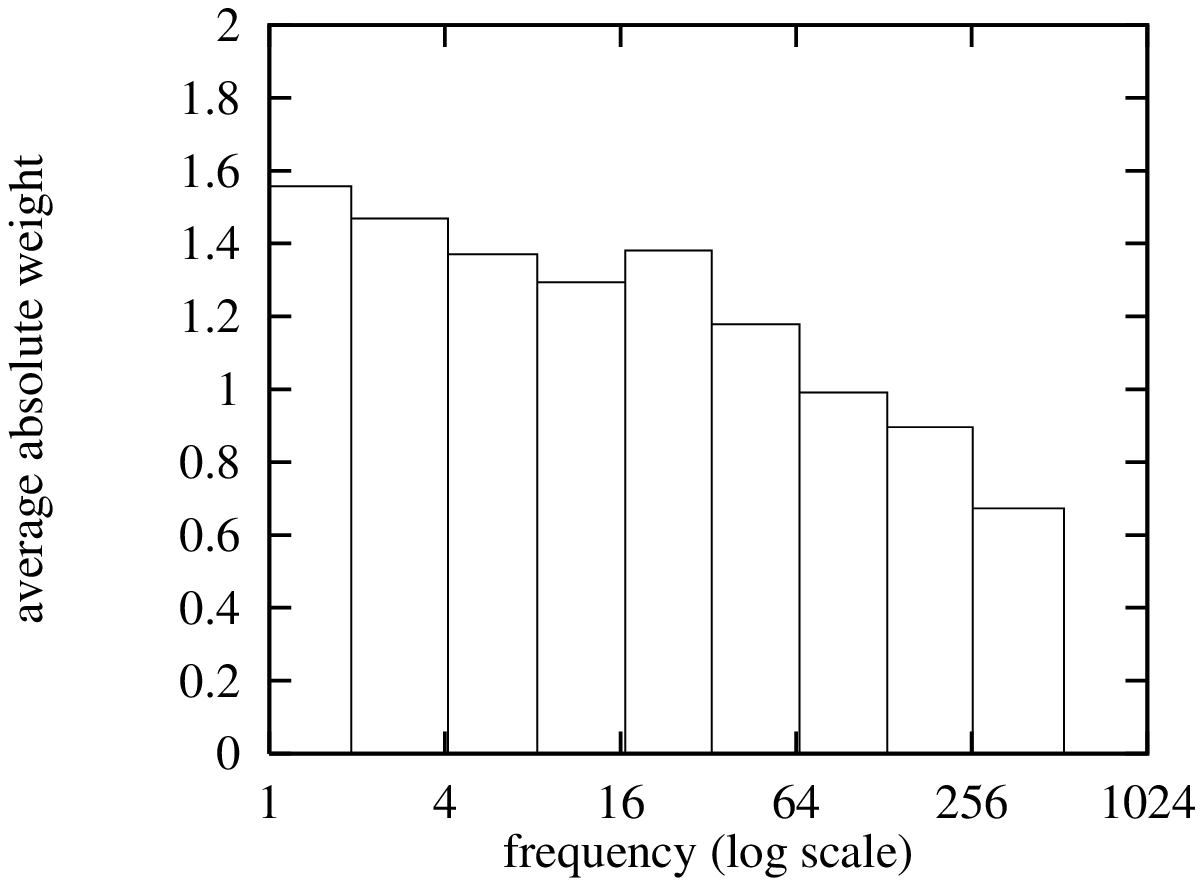,width=3.6in}
\caption{Weights plotted against frequency of input node occurring, for training corpus Tr 1 on Perceptron \label{ptron-wts}}
\end{center}
\end{figure}

After training we see that the distribution of weights in Hodyne
and Perceptron nets have
certain characteristics in common. In both cases there is a trend for
links on the least common input tuples to be more heavily weighted
than the more common: see Figures \ref{hod-wts} and \ref{ptron-wts}.

This characteristic distribution of weights can be understood when we examine 
the process by which the weights 
are adapted. Since we are processing negative as well as positive examples in 
the training stage, the movement of weights differs from that found with
positive probabilities alone.  Some very 
common tuples will appear frequently in both correct and incorrect strings. 
Consider a pair such as {\bf (start-of-sentence, open-subject)}. This will
often occur at the start of both grammatical and ungrammatical strings. The
result of the learning process is to push down the weights on the links to
both the ``yes" and the ``no" output nodes.

A significant number of nodes
represent those tuples
that have never occurred in a grammatical strings. A few 
nodes represent tuples that have
only occurred in grammatical strings (Tables~\ref{newts}~and~\ref{newts2}). 

\renewcommand{\baselinestretch}{1}
\normalsize

\normalsize

The relationship between frequency of occurrence and level of weight 
accompanies the decision to assess the correctness of a whole 
string, rather than the status of each element. Strings that 
are slightly wrong will include tuples that occur both in grammatical and 
ungrammatical sequences. As a consequence we see that the classifcation
decision can depend more on infrequently occurring tuples. In particular,
tuples that usually only occur in an ungrammatical string can have a 
significant influence on the classification task. 

\subsection{Direct update {\em versus} error minimization}

The use of direct update rather than error minimization  methods may also have
an effect on generalization. 
The traditional LMS measure can lead to situations 
where most input vectors are close to the target, while a few, or a single 
one, are distant. This may be desirable when noisy data is processed, but
not for our linguistic data, where we want precise fitting. We want to 
capture information from isolated examples.
 We want to classify strings that are ungrammatical in a single element as 
well as those that are grossly ungrammatical.  

\section{Conclusion}
\label{conclusion}

The original objective of this work was to see whether
the pattern matching capabilities of neural networks could be mobilised for
Natural Language Processing tasks. The working partial parser demonstrates
that they can be. 

Multi-layer Perceptrons were tried in the past, but it was found that single
layer networks were more effective, provided that the data was appropriately
converted to a higher order form. 
Some arguments against this approach centre
on the lack of a principled method to find the pre-processing, $\Phi$ function.
But though the methods of finding the initial $\Phi$ function are based on 
intuition, a close initial examination of the data can mean that this intuition
is founded on an understanding of the data characteristics.  In the case of 
the linguistic data support for the non-linear conversion function has 
come from information theoretic tools.
Though the setting of parameters is not data driven at the micro level, as in
a supervised learning environment,  the functions are chosen to 
capture some of the structure and invariances of the data. The development
of neural processors for an analysis of DNA sequences also illustrates this
\cite{lapedes}.

The analysis in the previous section illustrates the transparency of single 
layer networks, and indicates why they are such convenient tools. Compared 
to multi-layer Perceptrons, the parameters of the processor are 
more amenable to being interpreted. The Hodyne 
net in particular lends itself to further linguistic analysis. 
Furthermore, this approach has the advantage of fast two stage training.
The speed of training, measured in seconds, shows how quickly single layer
networks can fix their weights. Training times are hardly an issue.

A significant question of generalization ability is seen to relate to the 
ratio of testing to training data set size. Current work on generalization 
 has focused on  principled methods of determining training set 
size to ensure that the probability of generalization error is less than a
given bound.   Having implemented a preliminary prototype our work will 
continue with much larger corpora.

\small\normalsize
In the development of this technology we return to the fundamental question: 
how do we reconcile computational feasibility with empirical relevance? How
do we match what can
be done to what needs to be done? Firstly, in addressing the parsing
problem we start by  decomposing
the problem into computationally more tractable subtasks. Then we investigate
the data and 
devise a representation that enables the simplest effective processors
to be used. The guiding principles are to attack complexity by decomposing
the problem, and to adopt a reductionist approach in designing the
neural processors.

\renewcommand{\baselinestretch}{1}
\small
\bibliography{bib}
\bibliographystyle{unsrt}
\end{document}